\documentclass[10pt, conference, compsocconf]{IEEEtran}
\usepackage{graphicx} 
\usepackage{amsmath} 
\usepackage{amssymb}  
\usepackage{bm}
\usepackage{multirow}

\DeclareMathAlphabet{\mbf}{OT1}{ptm}{b}{n}
\def\Vec#1{\!\!\hbox{$#1$\kern-0.38em\lower0.85em\hbox{$\vec{}\,$}}\,}
\newcommand{\bcf}{\;\mbox{\boldmath ${\cal F}$\unboldmath}}

\graphicspath{{figs/}}

\begin{document}

\title{Relatively Lazy: Indoor-Outdoor Navigation Using Vision and GNSS}


\author{\IEEEauthorblockN{Benjamin Congram and Timothy D. Barfoot}
\IEEEauthorblockA{University of Toronto Institute for Aerospace Studies \\
Toronto, Canada\\
ben.congram@robotics.utias.utoronto.ca, tim.barfoot@utoronto.ca}
}

\maketitle

\begin{abstract}

Visual Teach and Repeat has shown relative navigation is a robust and efficient solution for autonomous vision-based path following in difficult environments.
Adding additional absolute sensors such as Global Navigation Satellite Systems (GNSS) has the potential to expand the domain of Visual Teach and Repeat to environments where the ability to visually localize is not guaranteed. 
Our method of lazy mapping and delaying estimation until a path-tracking error is needed avoids the need to estimate absolute states.
As a result, map optimization is not required and paths can be driven immediately after being taught.
We validate our approach on a real robot through an experiment in a joint indoor-outdoor environment comprising 3.5km of autonomous route repeating across a variety of lighting conditions.
We achieve smooth error signals throughout the runs despite large sections of dropout for each sensor.

\end{abstract}

\begin{IEEEkeywords}
field robotics; vision-based navigation; sensor fusion; localization;

\end{IEEEkeywords}

\IEEEpeerreviewmaketitle

\section{Introduction}\label{sec:introduction}

Autonomously driving the same route through an environment multiple times is a common task for mobile robots with applications to mining, warehouse robots, and guided tours.
Visual Teach and Repeat (VT\&R)~\cite{Furgale2010} has shown this task can be achieved using only a single stereo camera and one training example in extremely non-planar environments to few-centimetre-level accuracy and across reasonable appearance change~\cite{MacTavish2018}.
All state estimation in VT\&R is done in a relative framework.
Vehicle transformations and landmark positions are calculated with respect to neighbouring poses, not an absolute frame.
Global consistency in the map is not required.
As a result, VT\&R is computationally inexpensive and can handle large networks of paths.
No post-processing of the map is needed, meaning the robot can re-drive the path immediately after it is taught.

\begin{figure}[tb]
	\centering
	\includegraphics[width=1\linewidth]{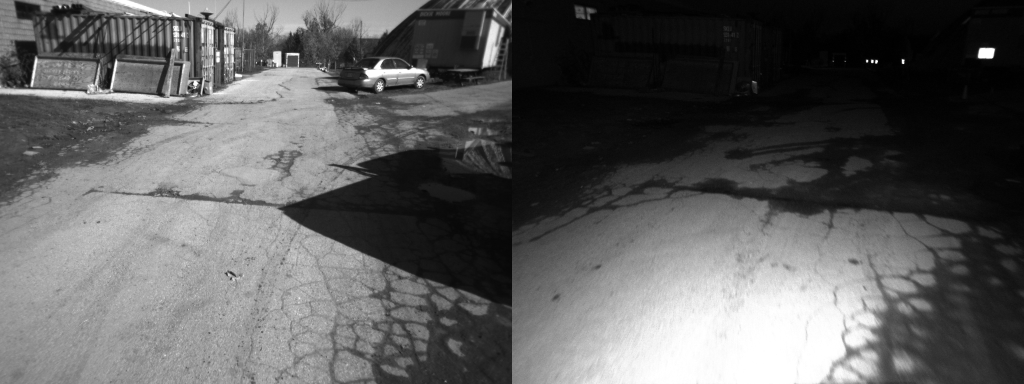}
	\caption{\footnotesize Comparison of images taken at the same location along a path highlighting the difficulty of visual localization in changing natural environments. Left: a keyframe from the daytime teach run. Right: an image from the nighttime repeat run (with headlights).}
	\label{fig:viz}
\end{figure}

\begin{figure}[tb]
	\centering
	\includegraphics[width=1\linewidth]{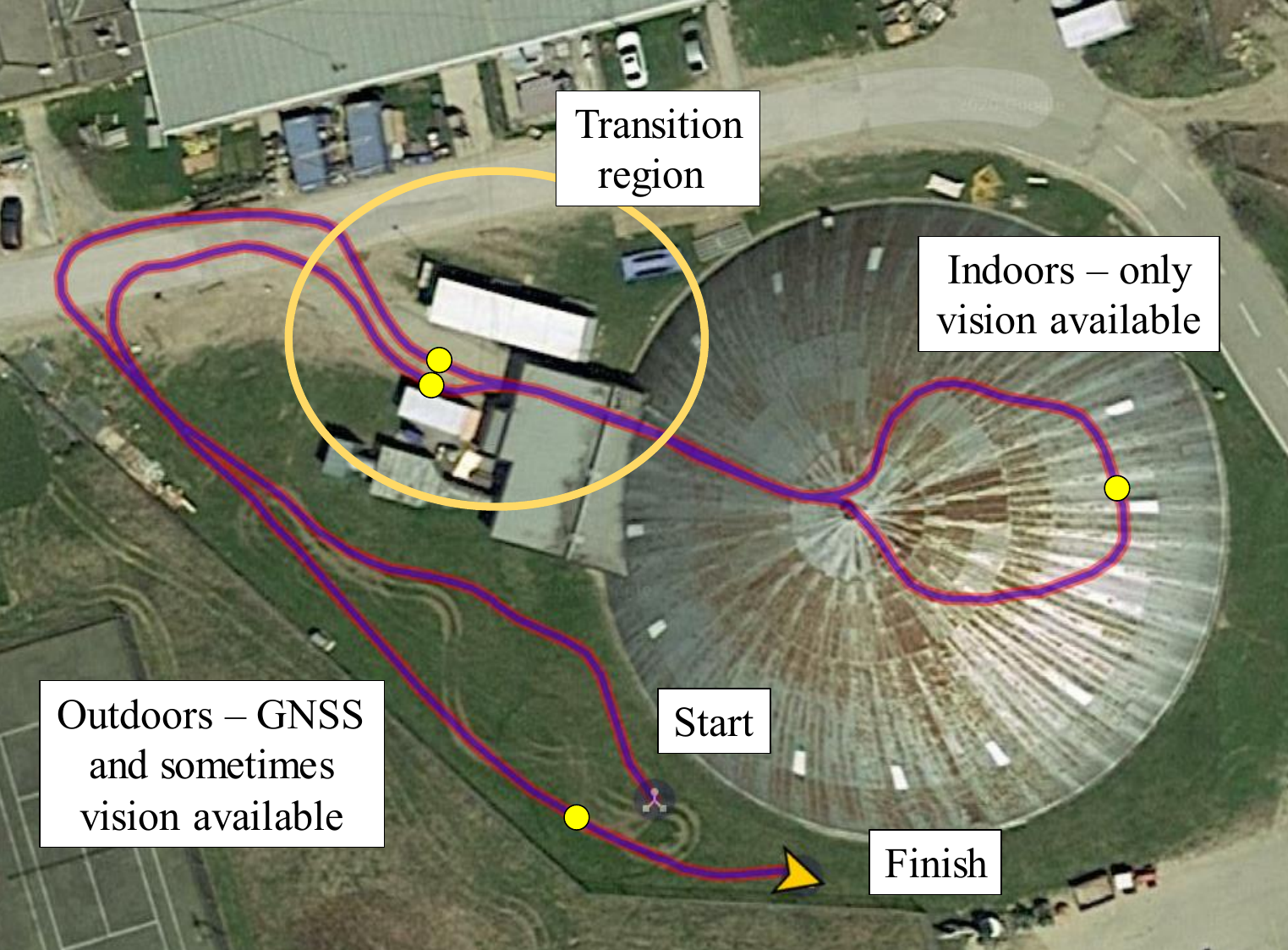}
	\caption{\footnotesize Overhead view of the approximate path driven in the experiments. Yellow dots denote the placement of key locations for independent ground-truth measurement of the path-tracking error. Inside the dome, the robot was forced to rely on only visual localization. Outside, the robot relied on mainly GNSS and used vision when available.}
	\label{fig:path_overhead}
\end{figure}

One challenge facing visual localization algorithms executed in unstructured environments, such as VT\&R, is appearance change.
Natural environments vary visually on both diurnal (e.g., lighting change) and seasonal time scales.
Figure~\ref{fig:viz} shows that this difficulty exists even when viewpoints can be matched nearly exactly.
Additional sensors may be used to achieve robust long-term navigation in all conditions.
Figure~\ref{fig:path_overhead} depicts a scenario where both Global Navigation Satellite Systems (GNSS) and vision could be utilized but the sensors must be integrated carefully to ensure smooth handoffs.

Absolute sensors such as GNSS present a challenge for relative navigation.
Unlike cameras, which observe in a local frame, GNSS provides observations in a global frame independent from the robot pose.
An accurate estimation of the global-to-local transformation is required to make use of the measurements in the path-tracking problem.
This is not guaranteed after a prolonged section of sensor dropout in which the robot relies on a form of dead-reckoning such as visual odometry (VO).

One option is to instead estimate all poses in a single global frame.
However, that would require a computationally-expensive batch map optimization to avoid jumps in the relative poses used in navigation.
Instead, we note that the key to the path-tracking problem is a good estimate of the path-tracking error --- the difference between the current and map states;
a good estimate of the states themselves (i.e., in a global frame) is not required.
This was the primary motivation to use a relative map in VT\&R\@.

Our method delays sensor fusion until the error estimation step.
We demonstrate this approach through a practical algorithm fusing GNSS and stereo imagery.
During the mapping phase, our solution simply logs GNSS observations and associates them with VO keyframes.
It makes no effort to force these to be consistent with the pose estimates from VO\@.
On repeat, a local window of GNSS measurements is used to estimate the orientation of the local frame with respect to the global frame and then rotate the error vector.
Our method adds little computational overhead to the already lightweight VT\&R algorithm.
It handles the transition between sensing modalities smoothly.
The regions of availability of each sensor do not need to be specified a priori.

In Section~\ref{subsec:relative-navigation}, we review related work on relative navigation while Section~\ref{subsec:vision-and-gnss} covers other algorithms fusing GNSS and vision.
Section~\ref{sec:methodology} describes our stereo-GNSS algorithm in detail.
Section~\ref{sec:experiments} introduces a field trial used for validation and the results are analyzed in Section~\ref{sec:results}.
Finally, we summarize the contribution and propose future work in Section~\ref{sec:conclusions-and-future-work}.

\section{Related Work}\label{sec:related-work}

\subsection{Relative Navigation}\label{subsec:relative-navigation}

Bundle adjustment (BA)~\cite{Brown1958} or batch simultaneous localization and mapping (SLAM), in which both vehicle poses and landmark positions are estimated from landmark observations, is the classic approach to large-scale localization and mapping.
However, computation time for the naive implementation scales quadratically with the number of landmarks~\cite{DurrantWhyte2006}, necessitating more efficient formulations.
Olson et al.~\cite{Olson2006} showed the benefit of using a relative-pose state space as opposed to global states in improving optimization performance.
Submapping~\cite{Chong1999}, \cite{Williams2001}, \cite{Marshall2008} has been used as a way to decouple computational complexity of the problem from map size.
Sibley et al.~\cite{Sibley2009} built on this with a completely relative formulation, doing away with any one privileged frame.
This allowed constant-time map updates even during loop closures.

VT\&R~\cite{Furgale2010} extended the idea of mapping on manifolds that are only required to be locally Euclidean to create a visual path-tracking system achieving high autonomy rates over many kilometres of highly non-planar terrain.
In VT\&R, a path is manually driven once as a single training example.
Then, the robot is able to autonomously repeat the path using only a single stereo camera, taking advantage of the deliberately consistent camera viewpoints.
The efficiency of the relative formulation allows the use of significantly larger factor graphs.
This led to experience-based navigation (EBN)~\cite{Churchill2013}, \cite{Linegar2015} as a method to increase robustness to appearance change.
The related multi-experience localization (MEL)~\cite{Paton2016} was added in VT\&R but with the ability to use landmarks from multiple experiences in the same metric localization problem and avoid drift from the original teach path over time.
Fast triaging of visual experiences~\cite{MacTavish2017b} is used to recall relevant landmarks in real-time.

Colour-constant image transformations have also been added to VT\&R to increase robustness to changing lighting conditions~\cite{Paton2015}.
VT\&R has been shown to generalize well to other sensors such as lidar~\cite{McManus2013} and monocular cameras~\cite{Clement2017}, as well as to other robot types such as unmanned aerial vehicles~\cite{Warren2019}.
In this work, we demonstrate a method to add a global sensor in GNSS to VT\&R while preserving the relative navigation formulation that is the key to its success.

\subsection{Vision and GNSS}\label{subsec:vision-and-gnss}

Vision and GNSS act as complementary sensors in many applications providing robustness via their independent failure modes.
Yu et al.~\cite{yu2019gps}  use visual-inertial odometry (VIO) to estimate local pose changes then use GNSS  and nonlinear optimization to bound the estimated drift.
They test on a unmanned surface vehicle with an omnidirectional stereo camera and show the benefit of added GNSS information over a short path.
Other works~\cite{schreiber2016vehicle}, \cite{kim2014high} develop similar methods combining VO and GNSS via an extended Kalman filter (EKF).
Qin et al.~\cite{Qin2019} offer a general algorithm to fuse a local sensor such as VO with a global sensor such as GNSS to estimate global poses
The local sensor provides high-rate local estimates of the path while pose-graph optimization is run with the global measurements to give low-rate estimates of the transformation to the absolute frame.

Fewer works attempt localization against a map using both vision and GNSS\@.
Choi et al.~\cite{choi2011localization} use a threshold on GNSS dilution of precision to switch between sensing modes.
Several works~\cite{li2018integration}, \cite{vishal2015accurate}, \cite{vysotska2015efficient} use GNSS signals only as a prior to simplify the image retrieval task.
Some then attempt metric localization using the retrieved images and the current camera frame.

Shi et al.~\cite{Shi2013} use GNSS observations offline to improve the global accuracy of visual SLAM\@.
Chen et al.~\cite{Chen2018} fuse vision and GNSS information by first estimating the frame transformation then solving a series of BA problems to generate globally consistent pose estimates.
Their work is most similar to ours.
However, they rely on estimating in an absolute frame, do not account for prolonged sensor dropout, and only test on a dataset with simulated GNSS\@.
Our work does not require a privileged frame or any post-processing of the map.
It handles transitions between sensing modalities smoothly and is robust to prolonged sections of sensor dropout.

\section{Methodology}\label{sec:methodology}

\subsection{Overview}\label{subsec:overview}

\begin{figure*}[tbhp]
	\vspace{2mm}
	\centering
	\fbox{\includegraphics[trim=0 2mm 0 2mm,clip,width=0.98\textwidth]{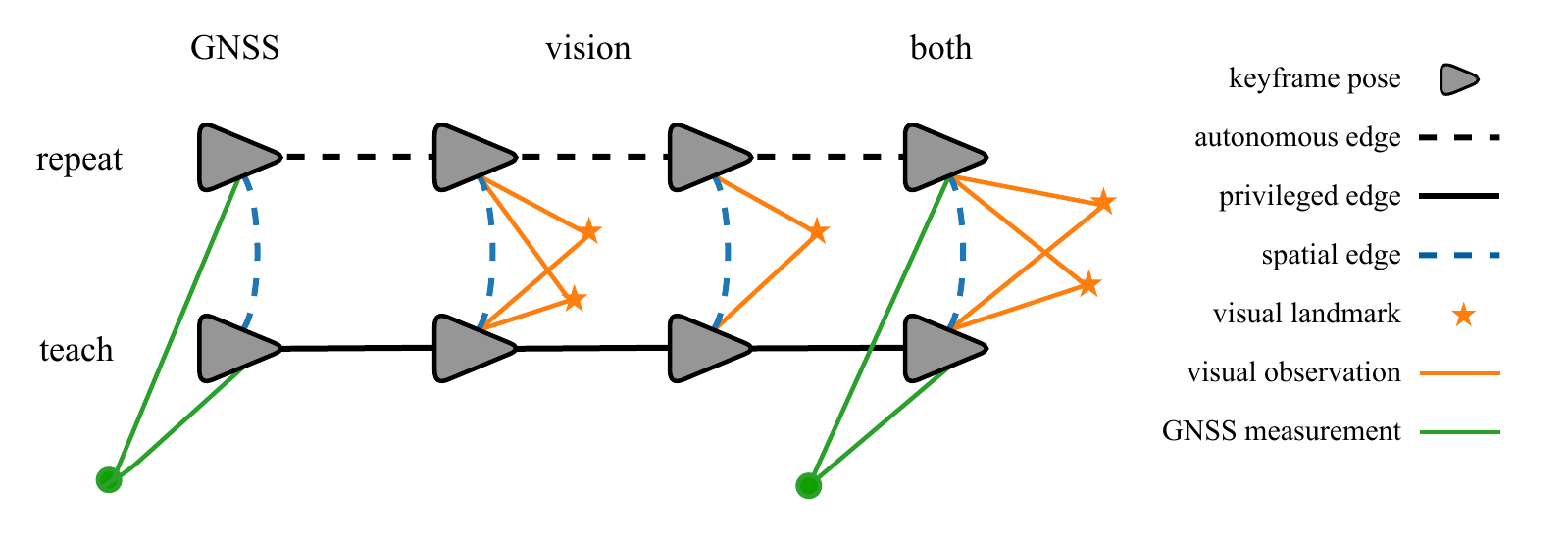}}
	\caption{\footnotesize Diagram illustrating how path-tracking error can be calculated using either vision or GNSS or a fusion of both. Privileged edges represent transformations calculated from the teach run VO while autonomous edges are calculated from repeat VO. Spatial edges come from localization. Path-tracking error is calculated for each sensor independently in the local frame. No single privileged frame is used, and the map does not need to be globally consistent.}
	\label{fig:triangles}
\end{figure*}

When path tracking, VT\&R uses stereo VO as a form of prediction and visual localization as a form of correction.
Our approach is to add GNSS as a second independent form of correction.
Each sensor can be used independently or fused when both are available as shown in Figure~\ref{fig:triangles}.
Other forms of prediction could be used in place of VO but this is beyond the scope of this work.
The VO pipeline we use is based on parallel tracking and mapping (PTAM)~\cite{klein2009parallel} in which motion estimates are calculated at frame rate while a windowed bundle adjustment is performed after each keyframe.
During the teach phase, the robot is manually driven and a pose graph is built through VO\@.
GNSS measurements are also logged and associated with each keyframe but no effort is made to make the poses from VO consistent with these observations.
In the repeat phase, a local window of GNSS observations from the current section of the map is recalled.
Together with a local window of the live repeat GNSS measurements, these are used to estimate a path-tracking error in the vehicle frame.
The details of this are described in Section~\ref{subsec:gnss-path-tracking-error}.
The GNSS error is then used to define the cost function:
\begin{equation}
	J_{\rm{GNSS}}= \frac{1}{2}\mbf{e}_{\rm{GNSS}}^T\bm{\Sigma}_{\rm{GNSS}}^{-1}\mbf{e}_{\rm{GNSS}}.\label{eq:gnss-cost}
\end{equation}
The GNSS cost is combined with the VT\&R vision cost function, $J_{\rm{vision}}$,  consisting of stereo landmark cost terms to form a nonlinear least squares optimization problem:
\begin{equation}
J= J_{\rm{vision}}+ J_{\rm{GNSS}}.\label{eq:combined-cost}
\end{equation}
It is solved with the Dogleg Gauss-Newton algorithm~\cite{Powell1964}.
The problem is well defined as long as at least one of vision or GNSS is available.
If neither is available at a given time, localization is not attempted and the robot relies on VO and a motion prior.

\subsection{GNSS Path-Tracking Error}\label{subsec:gnss-path-tracking-error}

To use a global sensor for vehicle navigation, an estimate of the vehicle's pose in a global frame is typically required.
Drift from dead-reckoning can produce arbitrarily poor pose estimates even if a full batch optimization is performed on the pose graph.
Because we estimate the GNSS path-tracking error independently, translational bias in the map is cancelled out when subtracting the live position from the map position.
Consequently, place-specific bias due to complex factors affecting GNSS such as multipath reflections does not reduce the performance of our algorithm.

After subtraction, the error vector is still in the global frame and needs to be rotated to the local frame.
We estimate the orientation of the local frame with respect to the global frame using linear regression on a local window of GNSS measurements.
The result is also reused to calculate the GNSS receiver position from the potentially noisy observations.
While vision performs 3D metric localization, we make a planar assumption in estimating our GNSS path-tracking error.
During logging, GNSS measurements are projected to Universal Transverse Mercator (UTM) coordinates resulting in data points with an $x$ and $y$ position and a timestamp.
On repeat, a small local window of these data points is recalled for both the teach and repeat runs.
For each window, least-squares regression is used to model the estimated position, $(\hat{x}, \hat{y})$, as a function of time, $t$:
\begin{equation}
\hat{x} = \bar{x} + \hat{\beta}_{1,x}(t-\bar{t}).\label{eq:x-lin-reg}
\end{equation}
\begin{equation}
\hat{y} = \bar{y} + \hat{\beta}_{1,y}(t-\bar{t}),\label{eq:y-lin-reg}
\end{equation}
$\hat{\beta}_{1,x}$ and $\hat{\beta}_{1,y}$ are the estimated linear regression slope parameters while $\bar{x}$, $\bar{y}$, and $\bar{t}$ denote the mean measurements within the current window.
This parameterization allows extrapolation to the current keyframe time for estimating $(x_m, y_m)$ and $(x_q, y_q)$, the map and repeat positions, respectively.
The model assumes a constant velocity for the vehicle --- a reasonable approximation for the small windows used (typically on the order of 1m).
More sophisticated models such as Gaussian processes could be employed but are outside the scope of this paper.
We also assume the vehicle velocity is parallel to the vehicle frame's $x$-axis (forwards), which is reasonable for non-holonomic robots.
The estimated heading, $\hat{\theta}$, is calculated using the slope parameter estimates from regression:
\begin{equation}
\hat{\theta} = \text{atan2}(\hat{\beta}_{1,y},\hat{\beta}_{1,x}).\label{eq:theta-reg}
\end{equation}
The heading estimate, $\hat{\theta}_{q0}$, for the live frame, $\Vec{\bcf}_q$, with respect to the UTM frame, $\Vec{\bcf}_0$, is used to generate the rotation matrix, $\hat{\mbf{C}}_{q0}$.
We calculate the error between teach and repeat positions in the UTM frame, $\mbf{r}_0^{mq}$, and rotate it to get an estimated position error in the live frame:
\begin{equation}
\hat{\mbf{r}}^{mq}_{q} = \hat{\mbf{C}}_{q0} \left( \hat{\mbf{r}}^{m0}_{0} - \hat{\mbf{r}}^{q0}_{0} \right) = \hat{\mbf{C}}_{q0} \begin{bmatrix}
\hat{x}_m - \hat{x}_q \\
\hat{y}_m - \hat{y}_q \\
0
\end{bmatrix}.\label{eq:r_mq}
\end{equation}
The result is assembled into the transformation matrix with a value of 0 set for the error in roll, pitch, and $z$-direction:
\begin{equation}
\hat{\mbf{T}}_{qm} = \begin{bmatrix}
\text{cos}\,\hat{\theta}_{qm} & \text{sin}\,\hat{\theta}_{qm} & 0 & \hat{r}^{mq}_{q,1} \\
-\text{sin}\,\hat{\theta}_{qm} & \text{cos}\,\hat{\theta}_{qm} & 0 & \hat{r}^{mq}_{q,2} \\
0 & 0 & 1 & 0 \\
0 & 0 & 0 & 1
\end{bmatrix}.\label{eq:T_qm}
\end{equation}

The state we are optimizing for in~\eqref{eq:combined-cost} is $\mbf{T}_{qm}$, the transformation between the vehicle pose at the current repeat frame and the vehicle pose at the nearest keyframe in the teach run.
Using the logarithmic map we define the $SE(3)$ error term~\cite{barfoot_ser17} for our GNSS error, $\mbf{e}_{\rm GNSS}$, as a function of our state and the transformation matrix calculated in~\eqref{eq:T_qm}: 
\begin{equation}
\mbf{e}_{\rm GNSS}= \text{ln}(\hat{\mbf{T}}_{qm}\mbf{T}_{qm}^{-1})^{\vee}.\label{eq:gnss-error}
\end{equation}
The covariance on our estimate, $\bm{\Sigma}_{\rm GNSS}$, can be calculated analytically from the linear regression.
However, this requires certain assumptions on the error properties of GNSS that may not be satisfied.
Instead, we find better results by estimating the uncertainties through empirical trials.
A liberal uncertainty is added in the roll, pitch, and $z$ direction as a weak prior.

\subsection{Outlier Rejection}\label{subsec:outlier-rejection}

GNSS measurements can be prone to outliers due to nonlinear effects and biases.
Our algorithm uses RANSAC~\cite{fischler1981random} prior to the linear regression to reject outliers within a window.
It then uses M-estimation after calculating the path-tracking error to account for estimates that disagree significantly with the prediction from VO or, if available, visual localization.
The  robust cost function  from the dynamic covariance scaling (DCS) family~\cite{Agarwal2013} that we utilize is
\begin{equation}
\rho(u) = \begin{cases}
\frac{1}{2}u^2 & u^2 \leq k^2 \\
\frac{2k^2 u^2}{k^2 + u^2} - \frac{1}{2}u^2 & u^2 > k^2
\end{cases}.\label{eq:dcs}
\end{equation}
A $k$-value of 2 is used in our experiments.
We find the robust cost function is rarely a factor when a strong GNSS fix is available but can be helpful when fewer satellites are seen.
It also prevents GNSS from degrading the performance of VT\&R when good vision is available.

\section{Experiments}\label{sec:experiments}

We conducted a set of experiments at the University of Toronto Institute for Aerospace Studies (UTIAS) to validate our approach.
The 350m path shown in Figure~\ref{fig:path_overhead} was manually driven at midday on October 23.
The path was intentionally chosen to feature a 120m section inside the UTIAS MarsDome midway through.
The terrain inside the MarsDome is a highly non-planar mix of gravel and dirt.
Outside, the environment is a mix of pavement and grass.
The path requires at least one transition from GNSS being available to unavailable and vice-versa.
As day turns to night, lighting conditions inside the MarsDome remain relatively consistent but appearance change impedes the robot's ability to visually localize outdoors as all repeats rely on the teach run as a single experience for localization;
MEL was deliberately not used in order to cause vision failures.

\begin{figure}[tb]
	\centering
	\includegraphics[trim=0 2mm 0 8mm,clip,width=1\linewidth]{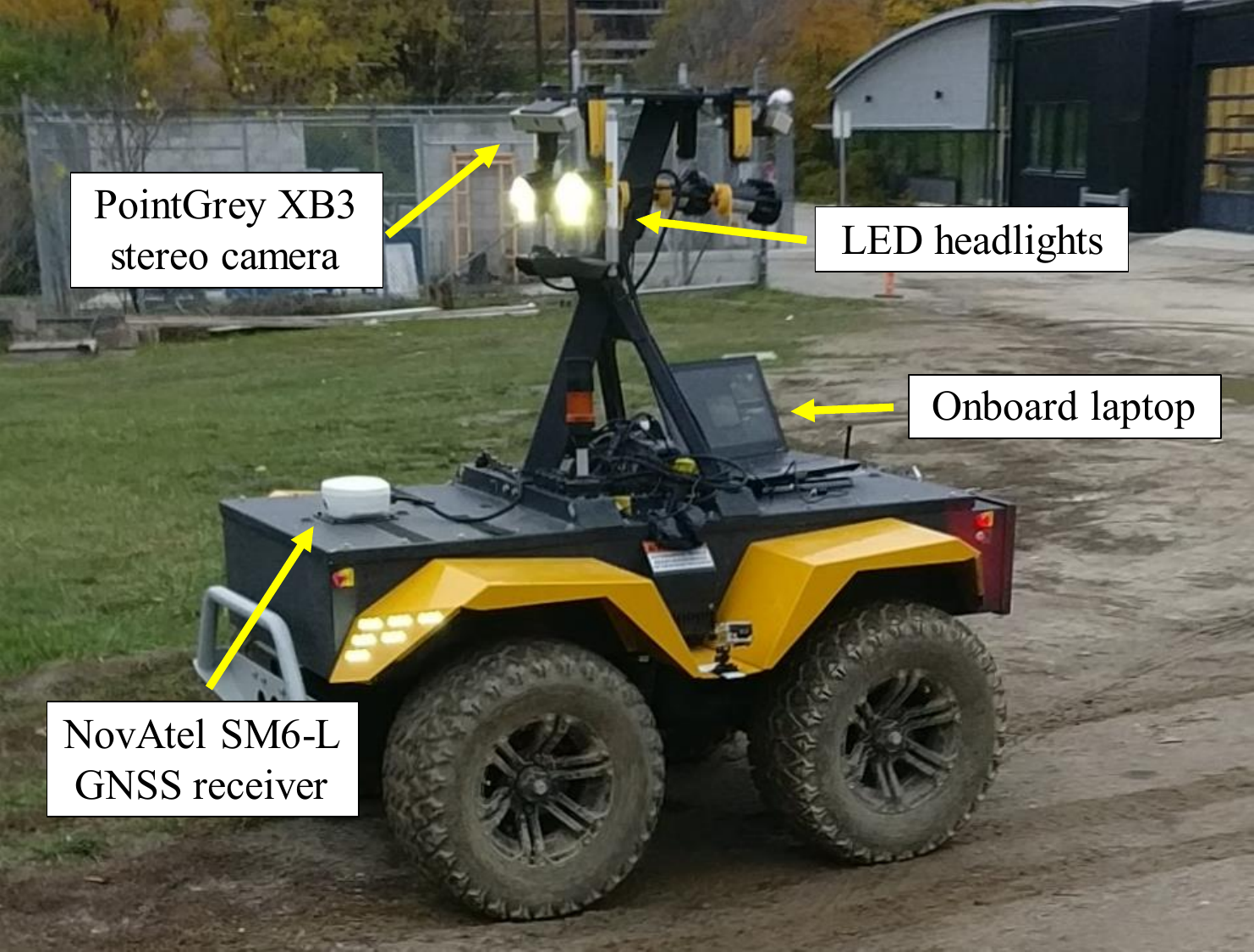}
	\caption{\footnotesize The Clearpath Grizzly Robotic Utility Vehicle used in the experiments seen autonomously driving the path.}
	\label{fig:grizzly}
\end{figure}

An initial repeat was conducted directly after the teach run to verify our ability to immediately re-drive paths.
Subsequent repeat runs were performed on the evening of October 26, beginning at 5:19pm.
They continued until after twilight with the final run beginning at 7:07pm.
Sunset occurred at 6:16pm.
The timing was chosen to demonstrate the algorithm's ability to both fuse vision and GNSS in the earlier repeats and transition to only GNSS navigation in the later repeats.

A Clearpath Robotics Grizzly Robotic Utility Vehicle, pictured in Figure~\ref{fig:grizzly}, was used for the experiments.
The Grizzly is equipped with a forward-facing Point Grey Research Bumblebee XB3 stereo camera featuring a 24cm baseline and a 66$^\circ$ horizontal field of view.
A NovAtel SMART6-L receiver is mounted on the front of the robot.
A second SMART6-L receiver serves as a stationary base station for Real-Time Kinetic (RTK) corrections.
The receivers are configured for Global Positioning System (GPS) satellites only in this experiment.
Sensor availability could be improved with additional GNSS constellations.
LED headlights are mounted below the stereo camera.
They are turned on for all runs to allow VO outdoors at nighttime while maintaining consistent conditions indoors throughout.
The algorithm runs on an onboard Lenovo laptop and interfaces with the Grizzly via Robot Operating System (ROS)~\cite{Quigley2009}.
All estimation is done in the vehicle frame with the fixed sensor-to-vehicle transformations explicitly handled for both the camera and GNSS receiver.

\begin{figure}[tb]
	\centering
	\vspace{2mm}
	\includegraphics[trim=0 4mm 0 4mm,clip,width=1\linewidth]{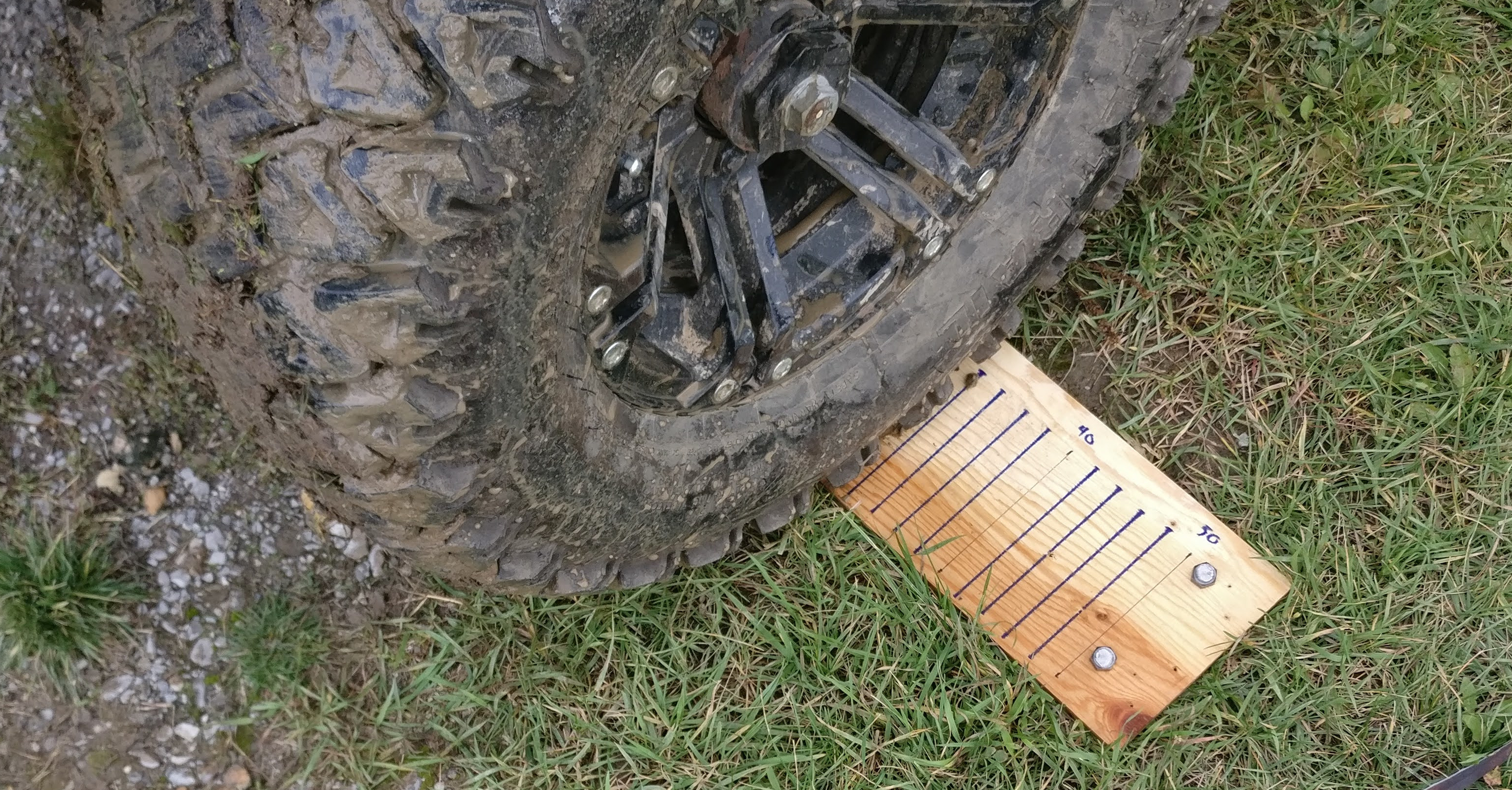}
	\caption{\footnotesize Jig used for measuring the ground-truth path-tracking error at several locations along the path. Lines are marked every 2cm and the board is affixed to the ground in the robot's path.}
	\label{fig:jig}
\end{figure}

To provide an additional measurement of path-tracking error independent from either sensor, jigs were placed at four locations along the path as indicated in Figure~\ref{fig:path_overhead}.
The jigs, seen in Figure~\ref{fig:jig}, consisted of a ruled board affixed to the ground over which the robot drives.
By comparing the tire positions of each repeat run to that of the teach run, a lateral and heading error could be estimated.
This simple method has the property of working well both indoors and outdoors.

\section{Results}\label{sec:results}

A total of 10 autonomous repeat runs of the experiment path were conducted totalling 3.5km of driving.
The first began immediately after the path was manually driven to validate the preservation of VT\&R's performance with good vision.
Immediate repeating remained easy and path tracking was highly accurate with the largest measured lateral error at the jigs just 2cm and all measured heading errors less than 1$^\circ$.
A summary of measured path-tracking errors for all repeats is provided in Table~\ref{tab:jig_errors}.

In general, the runs before sunset achieved good visual localization in most sections despite the different lighting conditions and the passing of several days.
As darkness set, visual localization was very difficult outdoors but remained relatively consistent indoors as expected.
All runs were successfully completed aside from the tenth in which the robot's headlights failed approximately 240m along the path in a section without a GNSS fix.
Lack of lighting led to failures in VO and the absence of GNSS increased the localization uncertainty beyond a threshold VT\&R deemed safe for continued driving so the robot stopped.
This unexpected event demonstrated the safe failure mode of VT\&R\@.

\begin{figure}[tb]
	\centering
	\includegraphics[width=1\linewidth]{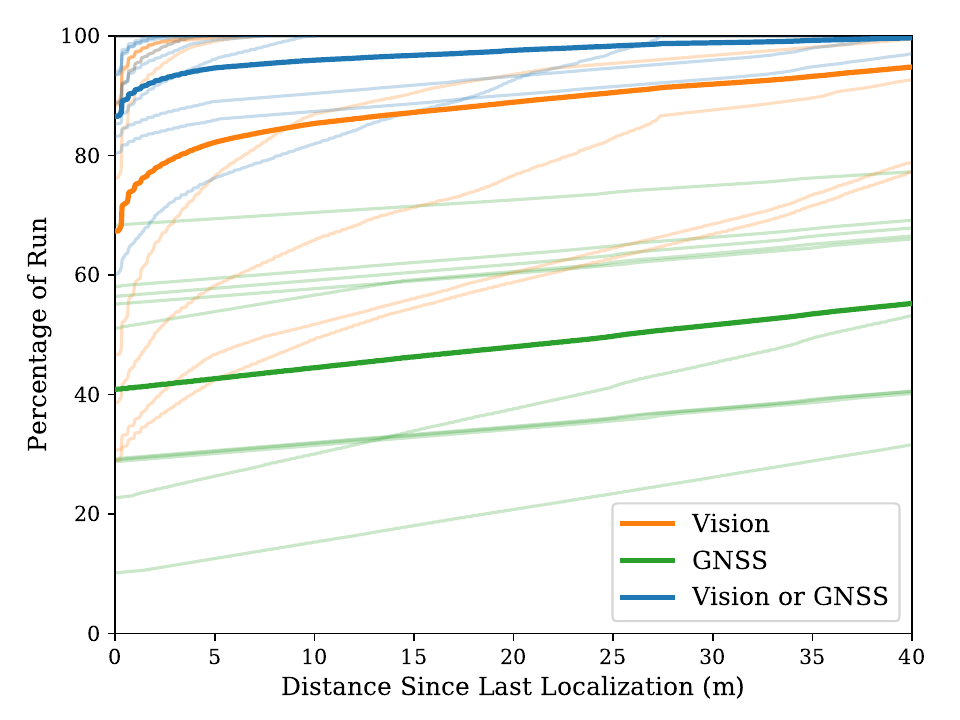}
	\caption{\footnotesize Plot showing the percentage of each run in which the robot has successfully localized within the last $x$ metres travelled with a particular sensor. Individual runs are shown in lighter colours while the mean is plotted in a darker colour. Top-left is better as it means the robot had to rely on dead-reckoning for shorter periods. Having both sensors available leads to a better chance at localization.}
	\label{fig:cumulative_dist}
\end{figure}

All runs experienced long sections of GNSS dropout due to the indoor driving.
Several runs experienced prolonged vision dropout for up to 90m.
The conditions across all repeats are summarized in Figure~\ref{fig:cumulative_dist}.
The results in this figure illustrate the importance of multiple sensors for robustness in difficult environments.
Our combined algorithm has to rely on VO alone for much shorter and less frequent sections.

The mean absolute path-tracking error across all runs, as measured at the four jig locations is just 2.9cm laterally and 0.4$^\circ$ in heading.
The measured error is a combination of both the localization error and the path-tracking controller error, as well as measurement error from the jig itself.
The metric localization error estimates from our algorithm are correlated with the measured error implying the mean absolute localization error is smaller than the total path-tracking error.
The only setting in which we see a noticeable inaccuracy is at the fourth measurement location during runs after sunset.
Here, the error is consistently 12--15cm suggesting that there may be some bias in the teach run GNSS measurements.
Another factor may be the manual driving of the teach path itself, which was slightly more erratic in this area of the path as the human pilot tried to target the centre of the jig.
That the error does not occur in earlier runs suggests our method is robust to GNSS bias when good vision is available and our sensors are weighted appropriately in (2), the localization cost function.

\begin{table}[tb]
\vspace{2mm}
\caption{Measured Error at Key Locations (see yellow dots in Fig. 2)}
\label{tab:jig_errors}
\begin{center}
\begin{tabular}{|c|c|c|c|c|c|}
\hline
\textit{\begin{tabular}[c]{@{}c@{}}Repeat \\Start Time\end{tabular}} & \textit{Error} & \textit{\begin{tabular}[c]{@{}c@{}}Entering \\ Dome\end{tabular}} & \textit{In Dome} & \textit{\begin{tabular}[c]{@{}c@{}}Exiting \\ Dome\end{tabular}} & \textit{Grass} \\ \hline
\multirow{2}{*}{1:54pm} & Lateral (m) & 0.020 & -0.005 & -0.010 & 0.005 \\ \cline{2-6}
& Yaw (deg) & 0.0 & -0.6 & -0.6 & -0.6 \\ \hline
\multirow{2}{*}{5:19pm} & Lateral (m) & 0.015 & -0.010 & -0.030 & -0.005 \\ \cline{2-6}
& Yaw (deg) & 0.6 & 0.0 & 0.6 & 0.6 \\ \hline
\multirow{2}{*}{5:30pm} & Lateral (m) & 0.005 & -0.005 & -0.020 & -0.005 \\ \cline{2-6}
& Yaw (deg) & 0.6 & 0.6 & 0.6 & 0.6 \\ \hline
\multirow{2}{*}{5:43pm} & Lateral (m) & 0.010 & -0.005 & -0.020 & -0.005 \\ \cline{2-6}
& Yaw (deg) & 0.0 & 0.6 & 0.0 & 0.6 \\ \hline
\multirow{2}{*}{5:58pm} & Lateral (m) & 0.010 & -0.005 & 0.000 & 0.000 \\ \cline{2-6}
& Yaw (deg) & 0.0 & 0.6 & 0.0 & 0.0 \\ \hline
\multirow{2}{*}{6:11pm} & Lateral (m) & 0.020 & -0.010 & -0.020 & 0.010 \\ \cline{2-6}
& Yaw (deg) & 0.0 & 0.0 & 0.6 & 0.0 \\ \hline
\multirow{2}{*}{6:23pm} & Lateral (m) & 0.040 & -0.010 & -0.030 & 0.125 \\ \cline{2-6}
& Yaw (deg) & 0.0 & 1.3 & 0.6 & 0.6 \\ \hline
\multirow{2}{*}{6:38pm} & Lateral (m) & 0.040 & -0.005 & -0.080 & 0.135 \\ \cline{2-6}
& Yaw (deg) & 0.0 & 0.6 & 0.6 & 0.6 \\ \hline
\multirow{2}{*}{6:51pm} & Lateral (m) & 0.065 & -0.010 & -0.040 & 0.155 \\ \cline{2-6}
& Yaw (deg) & 0.6 & 0.0 & 0.6 & 0.6 \\ \hline
\multirow{2}{*}{7:07pm} & Lateral (m) & 0.080 & 0.00 & -0.080 & -- \\ \cline{2-6}
& Yaw (deg) & 0.0 & 0.0 & 0.6 & -- \\ \hline
\end{tabular}
\end{center}
\end{table}

\begin{figure}[tbhp]
	\centering
	\fbox{\includegraphics[trim=0 4mm 0 4mm,clip,width=0.92\linewidth]{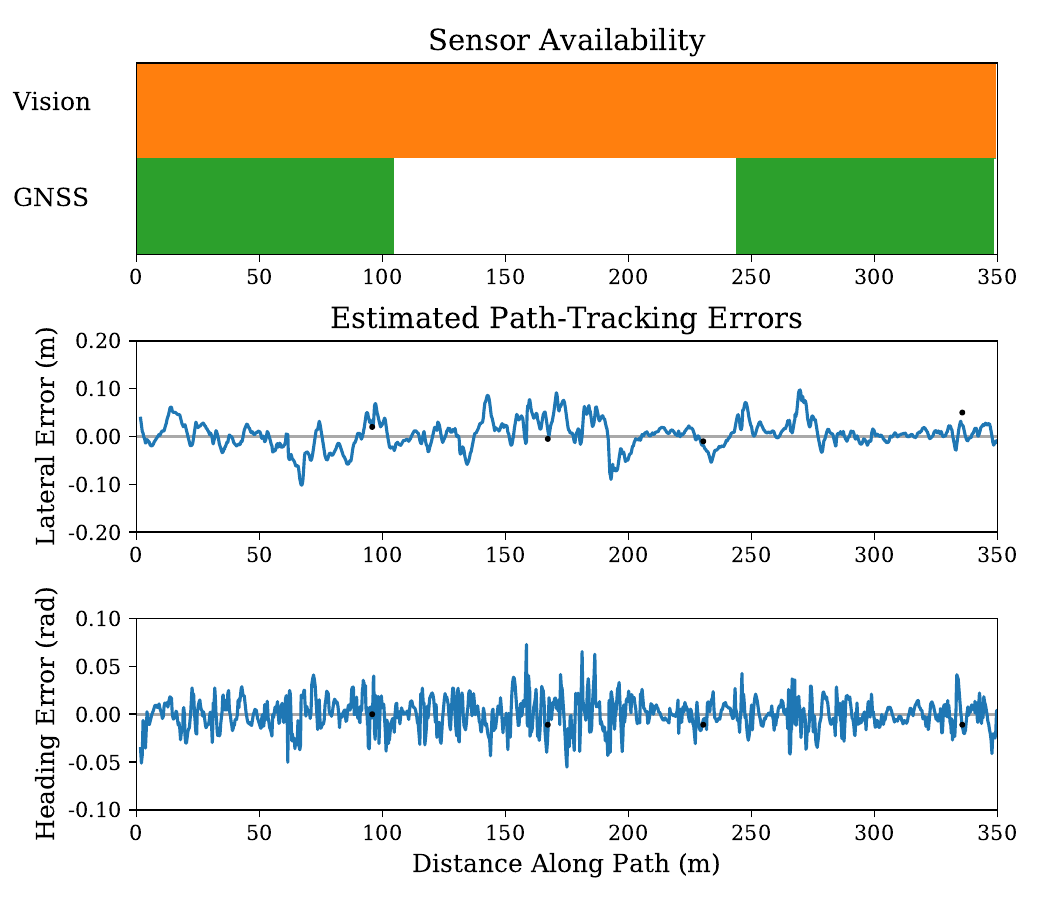}}
	\fbox{\includegraphics[trim=0 4mm 0 4mm,clip,width=0.92\linewidth]{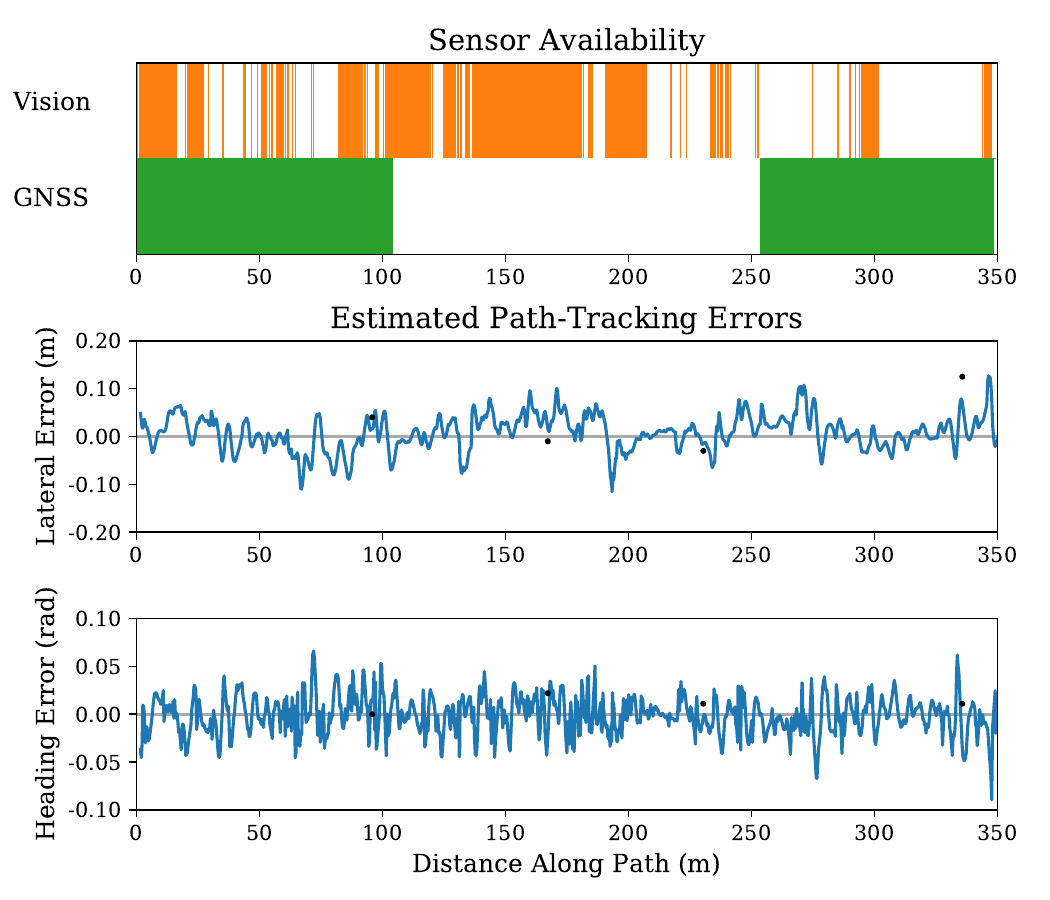}}
	\fbox{\includegraphics[trim=0 4mm 0 4mm,clip,width=0.92\linewidth]{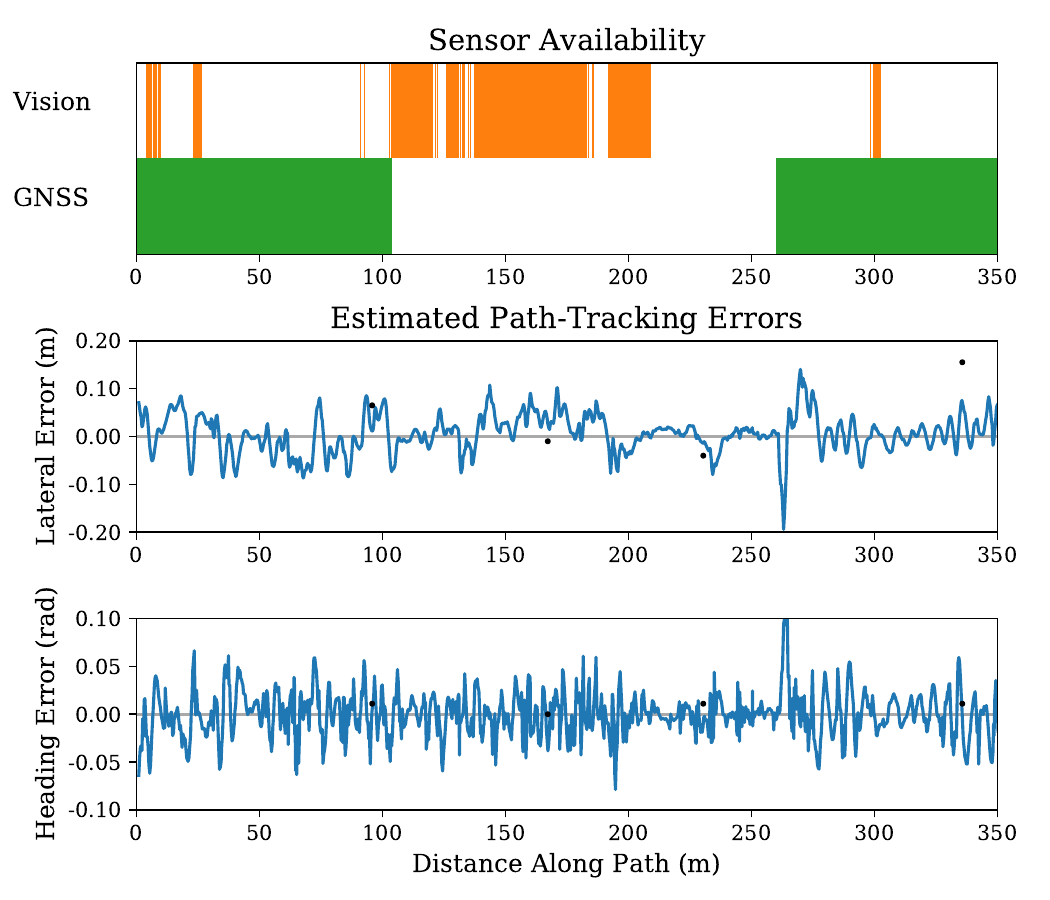}}
	\caption{\footnotesize Sensor availability versus estimated error for three repeats. Top: the 1:54pm run representing a best-case run. Middle: The 6:23pm run representing a typical repeat in the experiment. Bottom: the 6:51pm run representing the worst-case situation. The coloured bar represents when a sensor was used for localization. For example, the large gaps in the GNSS bars corresponds to the period when the robot was indoors. Black dots on the error plots correspond to the jig locations in Table~\ref{tab:jig_errors} where measurements of the true path-tracking error were taken.}
	\label{fig:run_errors}
\end{figure}

Equally important to path-tracking accuracy is the smoothness of the path following, especially in transition regions.
Figure~\ref{fig:run_errors} illustrates the relationship between the localization error estimated by the robot (not the true path-tracking error) and the availability of each sensor.
The first 100m of the path corresponds to outdoor navigation where GNSS is typically available.
It is followed by 120m indoors where the robot must rely on vision.
After exiting the MarsDome, the robot typically drives about 30m before regaining a GNSS fix for the final 100m outdoors.
The entrance to the MarsDome allows only a few centimetres of clearance for the Grizzly so smooth path following was not only a theoretical consideration but was of practical consequence.

Qualitatively, the path-tracking was very smooth throughout the experiment and no difference was noticed from changing sensor availability.
From the plots, we see the estimated localization error stays consistent throughout the transition zones.
The only spike that occurs is in the worst-case run when a GNSS fix is obtained after a prolonged section of VO reliance.
Crucially, this spike corresponds to the actual path-tracking error due to dead-reckoning, not a discontinuity in our map due to offset from the global frame.
The controller is able to smoothly utilize this new information to recentre the robot on the path.
We note the smaller estimated localization errors seen in the bottom plot of Figure~\ref{fig:run_errors} from approximately 210m to 260m are due to the lack of corrections from either sensor in this section, not because the true path-tracking accuracy is better.

Without our addition of GNSS, the later repeat runs would almost certainly have failed.
Consider the 6:51pm run whose sensor availability is shown in the bottom plot of Figure~\ref{fig:run_errors}.
During the first 100m, the robot has little success at visual localization meaning it would have to rely on VO nearly exclusively.
MacTavish at al.~\cite{MacTavish2017a} showed the mean drift rate for our VO pipeline in nighttime conditions is 2.38\% suggesting it is highly unlikely the robot could have safely navigated the dome entrance after 100m of dead reckoning.
In this scenario, the final third of the repeat would also need to rely exclusively on VO leading to expected final errors in excess of 3m compared to the few-centimetre-level errors we measured.
Our method allowed the robot to accurately and efficiently repeat the path in all experimental conditions.
Finally, we note that the small additional computation required to estimate GNSS error from the local windows of observations did not have a significant effect on the speed of the VT\&R algorithm and the runtime was well within the requirements to operate in real time.

\section{Conclusions and Future Work}\label{sec:conclusions-and-future-work}

This paper presented a robust system for path following utilizing both vision and GNSS\@.
Unlike related methods, we do not attempt to reconcile measurements from the two sensors into a single global coordinate frame.
By delaying sensor fusion until the path error is calculated, we avoid requiring a costly optimization for map updates, even after prolonged sensor dropout.
We validated our approach through an extensive field trial on a real robot.
We emphasize three key results: a) the system maintains high path-following accuracy on the order of centimetres, b) the vehicle was able to overcome long sections of dropout of one or both sensors, and c) there was no spike in error signal during the transitions between sensors due to frame offset and the vehicle continued to drive smoothly.

Several avenues exist for future work.
We performed our experiments using only RTK-corrected GNSS but the idea could transfer to less accurate GNSS setups provided the path-tracking error does not exceed the convergence region for visual localization.
This would be appropriate in many applications where the acceptable path-tracking error is larger outdoors in open space than indoors.
As a consequence of the relative framework, our algorithm only requires relative accuracy of GNSS so loop-closure techniques to increase single-point GNSS accuracy~\cite{Suzuki2020} could also be explored.

We assumed VO estimates are available for the prediction step but we could substitute other sensors or rely on GNSS alone in the event of total vision failure.
We are also not limited to GNSS; the `relatively lazy' approach holds for integrating other absolute sensors into relative navigation systems.
Finally, our approach need not be a replacement for multi-experience localization.
It could benefit from MEL by adding in GNSS later to sections where a fix was unavailable during the teach run and by using multiple experiences to average out GNSS noise.

We believe this work will be beneficial in a number of applications such as mining that involve both natural environments where appearance change is a factor and more structured but confined space where a GNSS signal is not guaranteed.
It could also have applications in marine robotics where visual features are sparse on open water but higher accuracy and robustness to satellite occlusion is required near shore.

\section*{Acknowledgment}
We would like to thank the Natural Sciences and Engineering Research Council of Canada (NSERC) and Clearpath Robotics for supporting this work.

\bibliographystyle{IEEEtran}
\bibliography{bib/refs}

\end{document}